\documentclass{article}

\usepackage{microtype}
\usepackage{graphicx}
\usepackage{subcaption}
\usepackage{booktabs} 

\usepackage{hyperref}

\usepackage{multirow}
\usepackage{makecell}
\usepackage[dvipsnames]{xcolor} 

\usepackage{algorithm}
\usepackage{algorithmic}

\usepackage[preprint]{icml2026}

\usepackage{amsmath}
\usepackage{amssymb}
\usepackage{mathtools}
\usepackage{amsthm}

\usepackage{listings}
\usepackage{tcolorbox}
\tcbset{
  systempromptbox/.style={
    colback=gray!10,
    colframe=black!40,
    boxrule=0.5pt,
    arc=2pt,
    left=6pt,
    right=6pt,
    top=4pt,
    bottom=4pt,
    fonttitle=\bfseries,
    title={System prompt}
  }
}

\usepackage[capitalize,noabbrev]{cleveref}

\theoremstyle{plain}

\theoremstyle{definition}

\theoremstyle{remark}

\usepackage[textsize=tiny]{todonotes}

\icmltitlerunning{Learning to Generate Formally Verifiable Step-by-Step Logic Reasoning via Structured Formal Intermediaries}

\begin{document}

\twocolumn[
  \icmltitle{Learning to Generate Formally Verifiable Step-by-Step Logic Reasoning\\
  via Structured Formal Intermediaries}



  \icmlsetsymbol{equal}{*}

  \begin{icmlauthorlist}
    \icmlauthor{Luoxin Chen}{yyy}
    \icmlauthor{Yichi Zhou}{comp}
    \icmlauthor{Huishuai Zhang}{yyy}
  \end{icmlauthorlist}

  \icmlaffiliation{yyy}{Wangxuan Institue of Computer Technology, Peking University.}
  \icmlaffiliation{comp}{ByteDance}

  \icmlcorrespondingauthor{Huishuai Zhang}{zhanghuishuai@pku.edu.cn}

  \icmlkeywords{Machine Learning, ICML}

  \vskip 0.3in
]



\printAffiliationsAndNotice{}  


\newcommand*{\fullname}[1]{{\emph{Process Reward over Structured Formal Intermediates}}}
\newcommand*{\shortname}[1]{{\emph{PRoSFI}}}
\begin{abstract}


Large language models (LLMs) have recently demonstrated impressive performance on complex, multi-step reasoning tasks, especially when post-trained with outcome-rewarded reinforcement learning \citep{guo2025deepseek}. However, it has been observed that outcome rewards often overlook flawed intermediate steps, leading to unreliable reasoning steps even when final answers are correct. 
To address this unreliable reasoning, we propose \shortname{} (\fullname{}), a novel reward method that enhances reasoning reliability without compromising accuracy. Instead of generating formal proofs directly, which is rarely accomplishable for a modest-sized (7B) model, the model outputs structured intermediate steps aligned with its natural language reasoning. Each step is then verified by a formal prover. Only fully validated reasoning chains receive high rewards. 
The integration of formal verification guides the model towards  generating step-by-step machine-checkable proofs, thereby yielding  more credible final answers. \shortname{} offers a simple and effective approach to training trustworthy reasoning models.


\end{abstract}

\section{Introduction}

Recent advances in techniques such as Chain-of-Thought (CoT) prompting and Reinforcement Learning (RL) have significantly enhanced the reasoning capabilities of large language models (LLMs). These methods have improved both the length and quality of model-generated reasoning chains, giving rise to the paradigm of test-time scaling \citep{snell2024scaling}. Representative systems include OpenAI o1~\citep{jaech2024openai} and DeepSeek‐R1~\citep{guo2025deepseek}, which set new records across multiple reasoning benchmarks.

Notably, DeepSeek‐R1-Zero~\citep{guo2025deepseek}, which employs  a rule‐based reward mechanism (outcome correctness plus output format score) to guide learning via the Group Relative Policy Optimization (GRPO) algorithm, successfully achieves strong multi‐step reasoning capabilities and exhibits emergent reflection behaviors. The simplicity and efficiency of this approach have inspired a series of follow‐up works, such as R1-Zero replications TinyZero~\citep{tinyzero} and Logical RL~\citep{xie2025logic},  and various GRPO variants~\citep{yu2025dapo,liu2025understanding}.

\begin{figure*}[t]
    \centering
    \includegraphics[width=\linewidth]{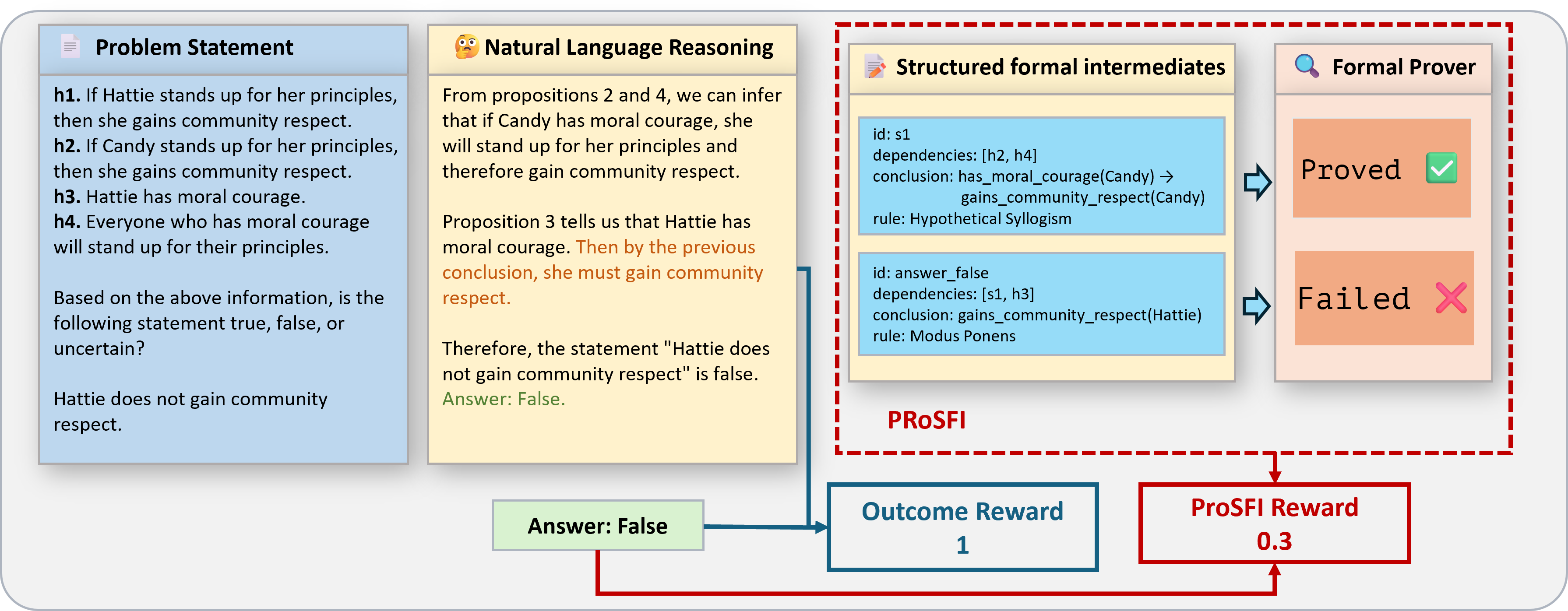}
    \caption{\textbf{Pipeline for \fullname{} and comparison with Outcome Reward.}  The process begins with a problem statement (left block) containing multiple propositions (h1–h4) related to individuals (Hattie and Candy). Then, a large language model (LLM) first generates natural language reasoning steps to solve the problem (middle block). The \textbf{Outcome Reward} approach then directly compares the generated answer with the ground truth, assigning a reward of 1 if they match. In contrast, \shortname{} introduces an additional step by generating structured formal intermediates that capture logical dependencies, inferred conclusions, and the application of formal inference rules such as \emph{Hypothetical Syllogism} and \emph{Modus Ponens}. These formal representations are then submitted to a \textbf{Formal Prover}, which verifies the logical soundness of the reasoning. In the illustrated example, although the generated answer (``Hattie does not gain community respect'' is false) is correct, the formal proof fails verification. As a result, \shortname{} assigns a low reward despite the correct outcome.}
    \label{fig:pipeline}
\end{figure*}

However, although these approaches greatly improve the final accuracy, they do not guarantee the correctness of the reasoning steps. In mathematical tasks, it has been observed that models may arrive at correct answers through invalid arguments \citep{uesato2022solving,lightman2024-lets-verify}. Empirical studies, including our own observations, show that RL guided by outcome-only rewards can degrade the faithfulness of reasoning chains \citep{chen2025reasoning}. Furthermore, in domains like proof generation, where final results cannot be automatically validated, outcome-based rewards are inapplicable. To ensure trustworthy reasoning, process-level supervision, not just outcome supervision, is necessary. 

Process-level supervision  provides feedback at each chain-of-thought step, yielding more precise training signals and producing models with more interpretable, reliable reasoning steps \citep{uesato2022solving}.  Nonetheless, collecting large-scale process-level feedback is challenging as human annotation at the granularity of individual proof steps is prohibitively expensive~\citep{lightman2024-lets-verify}. A natural solution is to leverage formal provers to verify intermediate reasoning steps. This approach requires access to formal codes, either directly generated  by language models or translated from  their natural language reasoning steps, which poses two key challenges.

First, models that excel in natural language reasoning often struggle to generate correct formal code. Training models to write formal proofs typically requires much larger model sizes and significant data distillation \citep{wang2025kimina,ren2025deepseek}. Moreover,  models optimized for formal proof generation often struggle with structure and format following \citep{xin2024deepseek}. Consequently, it is difficult to have a model capable of generating both natural language explanations and aligned formal proofs simultaneously.

Second, existing models that excel at formal proof generation typically are not designed to take natural language inputs, and there is a lack of reliable methods for translating natural language into formal code, as well as for evaluating the faithfulness of such translations. Furthermore, generating formal code is inherently more challenging for large language models than producing natural language reasoning. Accurate formal code generation often relies on test-time scaling techniques like Best-of-N (BoN), which can significantly increase computational costs particularly during RL post-training.


To address these challenges, we propose a novel and effective reward method: \fullname{} (\shortname{}, see Figure~\ref{fig:pipeline}).  Instead of generating full formal proofs directly, \shortname{} asks the model to output a series of reasoning blocks where each block contains a natural language explanation and a corresponding structured intermediate representation. This intermediate representation takes the form of a formatted array, usually in JSON or YAML, including an identifier, a logical rule, a derived conclusion (in formal language), and a list of dependencies (prior steps or premises). Then, formal propositions are extracted from each intermediate representation and passed to a formal prover for verification. The reasoning chain is considered trustworthy and receives the highest reward only if all steps are successfully verified.

Specifically, \shortname{} offers the following advantages. First, generating intermediate representations is easy. Existing models, even modest-sized (7B) language models, are typically well-equipped for generating structured intermediate representations, and in this case, they only need to produce formalized propositions for each intermediate step—without having to construct full formal proofs. This avoids the core challenges of formal proof construction. Second, structured formal intermediates facilitate formal verification. Specifically, formal provers are tasked with verifying individual propositions derived from the intermediate representation, rather than constructing complete proofs for the original problems. This ensures step-level alignment with the intermediate reasoning. Overall, the structured formal intermediates serve as an effective bridge between language models and formal provers. They reduce the burden on both sides: language models are freed from producing formal syntax, avoiding distribution drift, while provers benefit from simplified sub-problems defined in a detailed and verifiable format. 

To evaluate the effectiveness of our method, we conduct controlled experiments in a logical reasoning setting. We use the ProverQA dataset~\citep{qi2025large}, which is hard for current models and, as a synthetic dataset, provides high-quality formal representations of the problems. Through extensive experiments, we demonstrate that \shortname{} can significantly improve the reasoning soundness for complex logical problems. 
 We summarize our contributions as follows:
\begin{itemize}

  \item \textbf{Conceptually,} we introduce a new reinforcement learning paradigm with step-by-step rewards guided by formal provers, which, to our knowledge, has not been addressed in the literature. This shifts the focus from rewarding only final answers to ensuring the correctness of the entire reasoning process. We create a setup for training and evaluating verifiable reasoning models of a modest size (7B), greatly lowering the entry barrier for the community. 
  
  \item \textbf{Technically,} we propose \shortname{}, a framework that effectively bridges language models and formal provers. It enables  RL post-training by  incorporating  formal verification feedback into policy optimization.
  
  \item \textbf{Empirically,} we demonstrate that Qwen2.5-7B-Instruct, after \shortname{} RL post-training, can generate verifiable proofs on hard  first-order logic tasks, achieving this without relying on massive LLMs or extensive data distillation, marking the first such result at this scale. 
\end{itemize}

\section{Related Work}

\textbf{Reinforcement Learning Post-Training} has recently played a crucial role in training large language models, particularly for human preference alignment (RLHF) and enhancing complex reasoning abilities. Early work such as InstructGPT \citep{ouyang2022training} leveraged Proximal Policy Optimization (PPO) \citep{schulman2017proximal} to extract reward signals from human preference data, steering the model toward outputs more aligned with user expectations. However, the high computational cost of PPO on large-scale models motivated the development of more efficient alternatives.

One such alternative is Group Relative Policy Optimization (GRPO) \citep{shao2024deepseekmath}, which has been adopted in training DeepSeek-R1-Zero \citep{guo2025deepseek}.  DeepSeek-R1-Zero trains a pretrained model by using GRPO with only rule‐based rewards to achieve significant gains on complex mathematical reasoning tasks. Since then, many replication projects have become public \cite{tinyzero}.



\textbf{Logical Reasoning Datasets} serve as important benchmarks for evaluating the deductive and symbolic reasoning capabilities of LLMs. These datasets are typically constructed either by manual annotation or via automatic generation. Manually curated collections, such as Reclor \citep{yu2020reclor} and FOLIO \citep{han2022folio}, offer rich linguistic diversity and complex logical rules. 
Automatic dataset generation methods have been proposed. ProntoQA \citep{saparov2022language} and ProofWriter \citep{tafjord2020proofwriter} use predefined templates or rule‐based systems to generate large-scale reasoning problems with controllable complexity. 


Other synthetic logic‐reasoning datasets also include LogicPro~\citep{jiang2024logicpro}, which converts LeetCode program problems into narrative reasoning problems and leverages Python execution to extract and verify detailed intermediate solution traces; LogicBench~\citep{parmar2024logicbench}, a  natural language inference benchmark that covers propositional, first‐order, and non‐monotonic logics; and FLD~\citep{morishita2023learning,morishita2024enhancing}, which employs a set of first‐order logic rules to generate synthetic multistep proofs.



\textbf{Formal-Language-Augmented Reasoning Techniques} have recently attracted great attention to improve language models' rigor and reliability. In logical reasoning tasks, systems such as LINC \citep{olausson-etal-2023-linc} and Logic-LM \citep{pan-etal-2023-logic} treat the LLM as a semantic parser, translating natural language premises and conclusions into first‐order logic expressions. These expressions are then checked by external solvers like Prover9 \citep{mccune2005release} or Z3 \citep{de2008z3}. Logic-LM further incorporates a self‐correction loop, using solver error feedback to iteratively refine the symbolic translations until they pass validation. Similarly, Logic Agent \citep{liu2024logic} systematically applies propositional logic rules to structure and verify the model’s reasoning, enhancing both effectiveness and interpretability. Other related work on logical and program-aided reasoning includes \citep{pmlr-v202-gao23f,lightman2024-lets-verify,morishita2024-alt,olausson-etal-2023-linc,pan-etal-2023-logic,parmar-etal-2024-logicbench,qi2025-llm-provergen,ryu2025-clover,xu-etal-2024-faithful,yang2023-leandojo,zhou2024-dont-trust-verify}.

For automated theorem proving, OpenAI’s GPT-f system \citep{polu2020generative} first demonstrated that LLMs can generate formal mathematical proofs, achieving a 31\% success rate on the miniF2F benchmark \citep{zheng2022minif2f}. More recent advances along this direction include DeepSeek-Prover-V1.5 \citep{xin2024deepseek}, Kimina-Prover \citep{wang2025kimina}, DeepSeek-Prover-V2 \citep{ren2025deepseek}, and Seed-Prover \citep{chen2025seed}. Delta-Prover \citep{zhou2025solving} further leverages the sketching and reflection capabilities of general-purpose LLMs, solving 95.9\% of the MiniF2F problems through decomposition and iterative reflection.

In mathematical reasoning, recent work has begun to integrate provers into LLM reasoning: Don’t Trust: Verify (DTV) framework \citep{zhou2024dont} focuses on quantitative solutions by instructing LLMs to formalize their natural language steps into Isabelle \citep{paulson1994isabelle} proof scripts, which are then automatically verified. On benchmarks such as GSM8K \citep{cobbe2021training} and MATH \citep{hendrycks2021measuring}, DTV outperforms naive majority voting by filtering out unsound solution paths,  demonstrating that formal provers can supply valuable feedback for reasoning paths.

\textbf{Process Reward Models and Verification.} Early work \citep{uesato2022solving, lightman2024-lets-verify, wang2023math} train process reward models (PRMs) to provide fine-grained feedback for reasoning steps, achieving higher accuracy on GSM8K and MATH benchmarks. Recent research has shifted towards using LLMs as verifiers to detect errors in the reasoning process. For example, \cite{huang2025winning} employed a general-purpose model with a verification-refinement pipeline to achieve a gold medal in IMO 2025.


\section{Preliminaries on Dataset, Base Model,  Algorithms and Evaluation Metrics}\label{sec:preliminary}

\paragraph{Dataset Description} 
ProverQA~\citep{qi2025large} is a synthetic first-order logic reasoning benchmark generated by the ProverGen framework, which combines the linguistic diversity of LLMs with the rigor of a symbolic prover (Prover9~\citep{mccune2005release}). More details are provided in Appendix \ref{app:data}.


\paragraph{Model Description} The Qwen2.5-7B-Instruct model, an instruction-tuned variant of the Qwen2.5 family developed by Alibaba’s Qwen team, comprises 7 billion parameters and is released under the Apache License 2.0~\citep{qwen2.5}. It builds on the Qwen2.5 base LLM with further fine-tuning to follow instructions and engage in chat/dialogue. Specifically, it can produce  reliable structured outputs (e.g. JSON), parses tables, spreadsheets, and other semi-structured formats robustly.

\paragraph{Group Relative Policy Optimization (GRPO)} 
GRPO is a variant of Proximal Policy Optimization (PPO) tailored for training large language models (LLMs) without an explicit value (critic) model~\citep{shao2024deepseekmath}. The algorithm details can be found in Appendix~\ref{app:algorithm}.

\paragraph{Metrics}  
We introduce several metrics to evaluate the various facets of the model performance: \emph{Answer Correct Rate}, \emph{Reward Hit Rate} and \emph{Soundness}.





\newcommand{\rowstretch}{\rule{0pt}{3.5ex}}

\begin{table}[h]
\centering
\caption{Evaluation metrics and their definitions.}
\small
\begin{tabular}{lc}
\toprule
\textbf{Metric} & \textbf{Definition} \\
\midrule
Answer Correct Rate 
&\rowstretch $\dfrac{\# \text{Final Answer Correct}}{\# \text{Total Examples}}$ \\\midrule

Reward Hit Rate 
&\rowstretch $\dfrac{\# \text{Outputs with Max Reward}}{\# \text{Total Examples}}$ \\\midrule

Soundness 
&\rowstretch $\dfrac{\# \text{Answer Correct \& Path Valid}}{\# \text{Total Examples}}$ \\
\bottomrule
\end{tabular}
\end{table}

\emph{Answer Correct Rate} measures the percentage of examples where the model produces the correct final answer, regardless of the validity of the reasoning steps. 
\emph{Reward Hit Rate} reflects how often the model receives the maximum reward ($R=1$) under our reinforcement learning framework, that is, when the final answer is correct and all intermediate reasoning steps are formally verified. \emph{Soundness} assesses both answer correctness and reasoning path validity, which is defined as the proportion of examples where the final answer is correct and the reasoning path is logically valid.

We note that \emph{Reward Hit}, i.e., all structured  formal intermediates are verified, does not necessarily guarantee the \emph{Soundness} of the entire reasoning path. This is because the extracted formal propositions may not fully cover the entire reasoning path. In practice, rigorously assessing the soundness of a reasoning path would require extensive human evaluation, which is often infeasible at scale. As an alternative, we adopt state-of-the-art LLM-based evaluators, such as GPT-OSS-120B, to approximate \emph{Soundness}. In the experimental section, we will analyze potential biases introduced by these LLM-as-a-Judge metrics.


With the dataset, model, algorithm, and evaluation metrics defined, we now explore how to train LLMs to generate step-by-step reasoning that is formally verifiable.

\section{Method: \fullname{}}\label{sec:grpo-json}

To incorporate step-level verification into the reinforcement learning process, a natural approach is to (1) ask for the model to emit natural language reasoning step and formal proof script at the same time (e.g., in Lean 4~\citep{moura2021lean}), and (2) use a theorem prover to verify the correctness of each step. The verification outcomes can then be used as reward signals to distinguish valid reasoning chains from unsound ones.


\paragraph{However, direct generation of stepwise formal proofs fails in practice.}




We prompted the model to emit Lean 4 proof code immediately after natural-language reasoning steps, exploring multiple prompt templates and both Qwen2.5-7B-Instruct and Qwen2.5-Coder-7B-Instruct variants. However, these attempts yielded unsatisfactory results: only a small fraction of the generated Lean scripts compiled successfully, and those that did were overly reliant on generic tactics such as \textit{simp}. The proofs were typically brittle, poorly structured, and misaligned with the model's natural-language reasoning, and consequently ineffective for verifying the reasoning process.

We attribute this failure to two key challenges. First, modest-sized models (e.g., 7B) struggle to generate accurate and meaningful formal proofs. This makes it difficult for GRPO to bootstrap useful supervision from the generated formal codes, a finding consistent with recent progress in formal proof generation, which has primarily relied on much larger base models~\citep{wang2025kimina,ren2025deepseek}.


Second, the process of integrating formal verification into the GRPO rollout pipeline introduces significant computational overhead. The frequent calls to external formal provers disrupt GPU parallelism during sequence generation, drastically slowing down training and reducing scalability.

These limitations motivate our alternative solution: rather than having models directly generate full formal proofs, we introduce a structured intermediate representation that enables effective extraction of formal propositions and step-level verification with formal tools, which at the same time remaining within the capability of modest-sized language models. 





\subsection{The Design of \fullname{} }

To overcome the challenges encountered in directly generating formal proofs, we propose a novel approach: \fullname{} (\shortname{}), which introduces a structured JSON/YAML based intermediate representation as a bridge between natural language reasoning paths and formal verification. This format serves as a lightweight, verifiable scaffold that enables formal reward during reinforcement learning. Crucially, Qwen2.5-7B-Instruct, the base model, is well-suited for this task due to its strong formatting capabilities, particularly for structured JSON or YAML outputs~\citep{qwen2.5}.

\paragraph{Structured Formal Intermediates} In \shortname{}, the model is prompted to produce a sequence of reasoning steps in structured format, where each step explicitly encodes the structure of logical reasoning, which includes the following fields: \texttt{id}: a unique identifier for the step, used for reference;  \texttt{dependencies}: a list of premises this step depends on, which may include given conditions or previous steps;  \texttt{conclusion}: The new conclusion drawn in this step, expressed in a formal language; \texttt{rule}: The logical rule  applied to derive the conclusion.


To ensure each step remains atomic, we impose a constraint that the number of dependencies per step must be fewer than 5 (tunable). Each step can then be converted into a sub-problem for verification. Moreover, we also require that the final step's conclusion match the target proposition, and that its identifier be among the provided answer candidates.





This structured format allows natural language reasoning to be decomposed  into discrete, formalizable sub-problems. For the first-order logic (FOL) tasks considered in this paper, which are relatively simple and well-structured, each proposition can typically be verified using lightweight tactics in Lean (e.g., \texttt{aesop, simp}), or through external automated solvers such as {Z3}~\citep{de2008z3} or {Prover9}~\citep{mccune2005release}. For more complex settings, we optionally delegate verification to powerful formal models such as {DeepSeek-Prover-V1.5}~\citep{xin2024deepseek}.  These configurations  can be adapted with different formal formats or verification tools for different tasks and domains.

\paragraph{Reward Design} If all steps pass formal verification, the entire reasoning path will be assigned a high reward $R=1$. Otherwise, a failure of any individual step flags a potential flaw in the reasoning process. Based on this framework, we define the reward function as follows:
\vspace{-2mm}
\[
R = 
\begin{cases}
  1.0, & \text{Answer correct and all steps verified,} \\
  0.3, & \text{Answer correct, but some steps not verified,} \\
  0.1, & \text{Output format correct, but answer  incorrect,} \\
  0.0, & \text{Output format incorrect or other failure cases.}
\end{cases}
\]
\vspace{-2mm}

This reward design of \shortname{} exploits the rigor of formal provers while avoiding the difficulties of generating full formal proofs, resulting in more stable, fine-grained, and structure-aligned supervision signals, which are crucial for effective reinforcement learning. Moreover, \shortname{} avoids the efficiency bottlenecks typically associated with integrating formal provers into the rollout phase of RL training. The model generates the full natural language reasoning and the  structured  formal intermediates in a single, uninterrupted forward pass. Formal verification is performed after generation, allowing it to be fully decoupled from the autoregressive process. This decoupling not only improves training throughput by enabling batch-level parallelism but also allows verification tools to run asynchronously, maximizing efficiency.


We next present empirical results to demonstrate that \shortname{} can effectively improve the soundness of LLM reasoning paths.

\section{Experimental Results}

\subsection{Training and Testing Setup}

\paragraph{Training Setup}
As described in Section~\ref{sec:preliminary}, our experiments are conducted on the ProverQA dataset, a synthetic logical reasoning benchmark designed with controlled difficulty levels and well-defined ground-truth labels (True/False/Uncertain). This makes it particularly well-suited for evaluating models in a controlled reinforcement learning (RL) setting. To address the difficulty imbalance within the ProverQA dataset, we excluded instances labeled as Uncertain. The remaining data were then partitioned into training and test sets for each difficulty level (easy, medium, and hard), yielding a total of 769 instances for the training set.

We adopt Qwen2.5-7B-Instruct as our base model. This model offers a good choice for strong performance given the model size under our hardware constraints. Moreover, it has demonstrated superior responsiveness to RL-based post-training compared to other open-source alternatives such as the LLaMA series~\cite{shao2025spurious, zeng2025simplerl}.

We use the GRPO algorithm with a learning rate of $3\times10^{-7}$ and batch size of 64. For each training iteration, GRPO samples 8 responses, with the maximum sequence length set to 5120 tokens. The KL coefficient in the GRPO loss function is set to 0.001. To enhance training stability, we implement the clip-higher strategy from the Decoupled Advantage Policy Optimization (DAPO) algorithm~\citep{yu2025dapo}, whose details are in Appendix~\ref{app:algorithm}.

All experiments are conducted on a compute node equipped with 8× Hopper GPUs, running on Debian GNU/Linux 12 with CUDA 12.4, PyTorch 2.6.0, vLLM 0.8.3, and Verl 0.4.0. Each training run consists of 960 iterations, which takes approximately 14 hours to complete.



\paragraph{Testing Setup}


Besides the test splits of difficulty levels in the ProverQA dataset, we introduce a more challenging out-of-distribution (OOD) test set, named ProverQA-Extra. This set consists of 128 synthesized examples generated from the ProverQA repository using Qwen2.5-32B-Instruct as the background story generation model. To reduce randomness, we sampled 16 responses for each test data instance and took the average of their results.




\subsection{Outcome Reward Leads to High Answer Correct  Rate But Low Soundness} \label{sec:initial}


We employ GRPO with a reward scheme similar to that of DeepSeek‐R1‐Zero to replicate its training methodology on ProverQA. This serves a baseline for the evaluation of reasoning path soundness.

\paragraph{Reward Design}

Following the DeepSeek‐R1‐Zero pipeline, the model must emit its chain of thought wrapped in \texttt{<think>} … \texttt{</think>} tags before the final answer, referred to as \emph{format correct}. We define the outcome-based reward \(R\) as: 
\[
R = 
\begin{cases}
  1.0, & \text{format correct and answer correct,}\\
  0.1, & \text{format correct but answer incorrect,}\\
  0.0, & \text{format incorrect.}
\end{cases}
\]
This reward encourages adherence to the output format and correctness of the final answer.

\paragraph{Prompt Design} The system prompt, adapted from DeepSeek‐R1‐Zero, is provided in Appendix~\ref{app:OutcomePrompt}.

The user prompt starts with the original dataset description, then presents the problem in formal language with clearly indexed premises (see Figure \ref{fig:pipeline}). We refer to this setup as \emph{Outcome Reward with CoT} or \emph{Outcome-CoT} for short. The full user prompt is presented in Appendix~\ref{app:OutcomePrompt}. 


\paragraph{Results: GRPO with \emph{Outcome-CoT} significantly boosts the Answer Correct Rate but achieves rather low Soundness.}
 Specifically, in our evaluation on the ProverQA-Hard subset, GRPO with \emph{Outcome-CoT} substantially improves the Answer Correct Rate, rising from 46.0\% with Qwen2.5-7B-Instruct to 91.31\%. This confirms the successful replication of the R1-Zero training procedure. However, the GPT Soundness score remains low, improving only from 9.0\% to 21.97\%. This indicates that although the final answers are often correct, the generated reasoning paths frequently contain flawed or invalid inference steps.

\begin{table*}[ht]
\centering
\caption{Results on ProverQA Hard and Extra. GPT Soundness values include the difference of each method versus the Outcome-CoT baseline.}
\renewcommand{\arraystretch}{1.5} 
\begin{tabular}{llcc}
\toprule
Method & Metric (\%) & Hard & Extra \\
\midrule
\multirow{2}{*}{Outcome-CoT} 
 & Answer Correct Rate & 91.31 & 86.96 \\
 & GPT Soundness & 21.97 & 11.87 \\
\midrule
\multirow{2}{*}{Direct Lean Generation} 
 & Answer Correct Rate & 92.29 & 83.15 \\
 & GPT Soundness & 
   \makecell{33.11 {\scriptsize\color{blue}{$\uparrow 11.1$}}} & 
   \makecell{10.45 {\scriptsize\color{red}{$\downarrow -1.4$}}} \\
\midrule
\multirow{2}{*}{ProSFI} 
 & Answer Correct Rate & 92.97 & 91.94 \\
 & GPT Soundness & 
   \makecell{76.07 {\scriptsize\color{blue}{$\uparrow 54.1$}}} & 
   \makecell{56.10 {\scriptsize\color{blue}{$\uparrow 44.2$}}} \\
\bottomrule
\end{tabular}
\label{tab:ProverQA-combined}
\end{table*}

\subsection{Direct Lean Generation Fails to Provide Reliable Stepwise Verification}

As a direct baseline for integrating formal verification into reinforcement learning, we evaluated an approach in which the model is required to emit complete Lean~4 proof scripts (rather than the structured intermediates introduced in \shortname{}). This baseline was designed to isolate the effect of output modality: reward schedule and training protocol were kept the same as in the structured-intermediate condition, and the only difference was that the model produced full formal proofs immediately following each natural-language reasoning trace.

Empirically, this direct-generation strategy failed to provide a useful stepwise verification signal for GRPO. Compile-and-verify rates were low: the Reward Hit Rate on ProverQA-Hard was 13.77\% and on ProverQA-Extra was 4.35\%. Instances that did compile were vanishingly rare, and when compilation succeeded the resulting proofs typically reduced to a single opaque tactic (e.g. a one-line \texttt{simp} invocation) that bore little structural correspondence to the model's natural-language chain of thought. Consequently, the direct-Lean baseline produced almost no reliable, fine-grained supervision: both final-answer accuracy and measured reasoning soundness decreased relative to the Outcome-CoT baseline, as shown in Table~\ref{tab:ProverQA-combined}.

\subsection{\shortname{} Learning with ProverQA}

\paragraph{Results: Structured formal rewards improve final accuracy with significantly more credible reasoning path.}

After GRPO training with stepwise formal rewards, we achieve comparable final-answer accuracy to the GRPO training with only final-answer reward on ProverQA, as shown in Table~\ref{tab:ProverQA-combined}. This confirms that our stepwise rewards can effectively be combined into R1-Zero reward with high reasoning accuracy. We can also see that stepwise formal rewards can significantly improve the credibility of reasoning paths in comparison with final-answer only reward in Section~\ref{sec:initial}. Moreover, the results on ProverQA-Extra also demonstrates \shortname{} achieves strong out-of-distribution (OOD) performance on logical reasoning questions of previously unseen difficulty.

\subsection{Test-Time Scaling Behaviors of \shortname{}}

A critical advantage of \shortname{} is its strong \emph{test-time scaling} behavior, made possible through integration with the \emph{Don't Trust; Verify} (DTV) method~\citep{zhou2024dont}. DTV is designed to enhance both accuracy and reasoning reliability by sampling multiple reasoning paths and employing a lightweight verifier to assess the logical validity of each output. If any candidates are successfully verified, one of them is selected; otherwise, the method defaults to majority voting. \shortname{} is particularly well-suited for integration with DTV due to its structured intermediate representations, which enable reliable formal verification. In contrast, approaches that rely solely on natural language reasoning lack the structured outputs required for effective collaboration with DTV, limiting their ability to benefit from test-time verification.

\begin{figure}[ht]
    \centering
    \includegraphics[width=\linewidth]{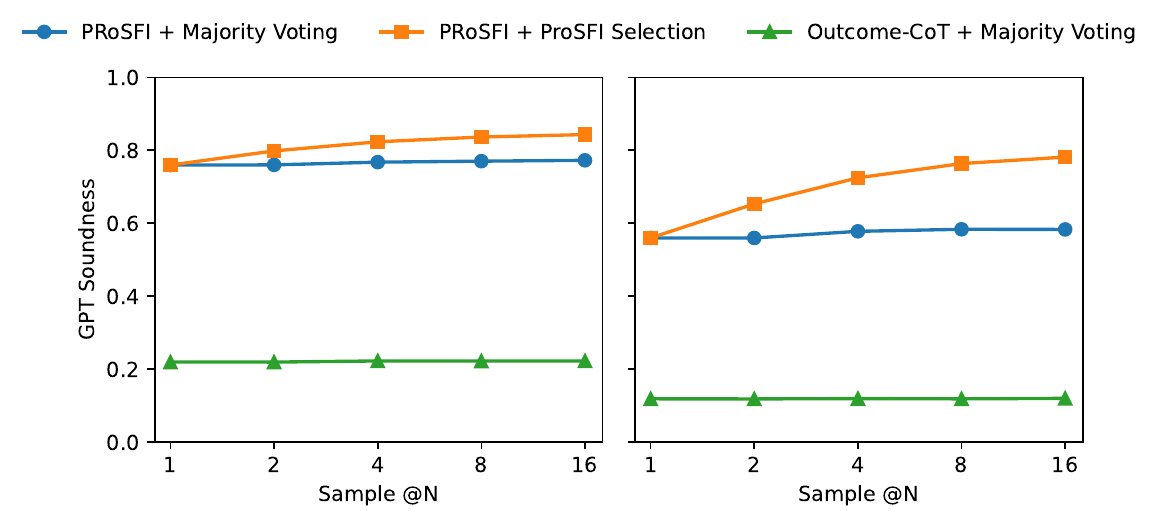}
    \caption{\textbf{Left:} GPT Soundness on  \texttt{ProverQA-hard} subset. 
    \textbf{Right:} GPT Soundness on  \texttt{ProverQA-extra} subset. 
    As the number of sampled paths increases, \shortname{} with \shortname{} selection consistently improves,  while \texttt{Outcome-CoT} cannot benefit due to translation limits.}
    \label{fig:dtv-scale}
  \end{figure}

In our experiments, we apply the DTV method to the outputs generated by \shortname{}. As shown in Figure~\ref{fig:dtv-scale}, increasing the number of sampled reasoning paths consistently improves performance on the Soundness metric, demonstrating \shortname{}'s robust and scalable behavior at test time. This trend holds across both the \texttt{ProverQA-Hard} and \texttt{ProverQA-Extra} subsets. In contrast, methods such as \texttt{Outcome-CoT} fail to benefit from additional sampling, as majority voting alone is insufficient to discern the soundness of the reasoning paths and cannot effectively filter out logically invalid reasoning.



  \begin{figure}[H]
      \centering
    \includegraphics[width=\linewidth]{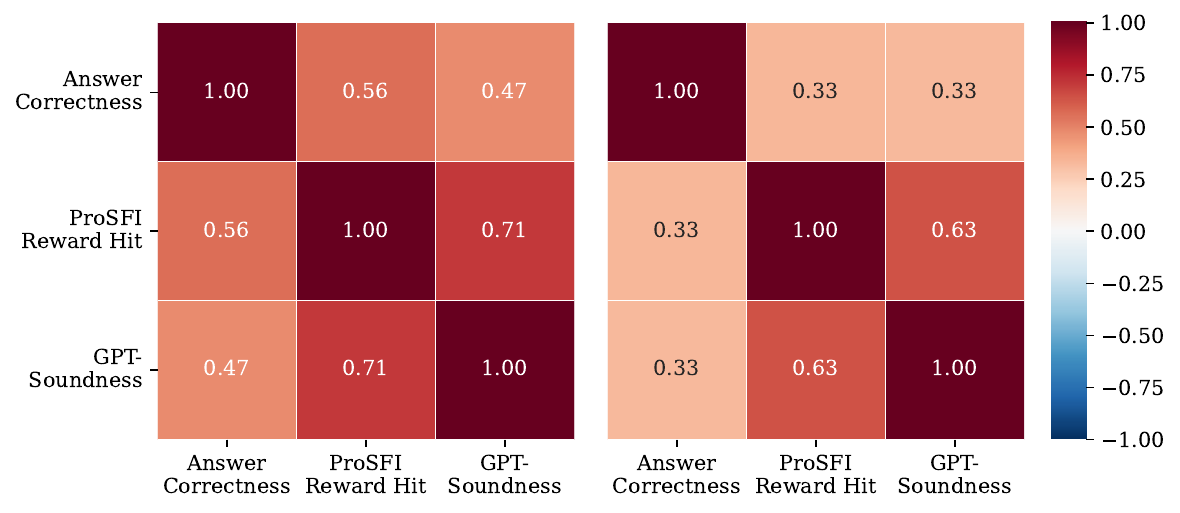}
    \caption{\textbf{Left:} Metric correlation heatmap on \texttt{ProverQA-Hard}. 
    \textbf{Right:} Metric correlation heatmap on \texttt{ProverQA-Extra}. 
    Compared to Answer Correctness, \shortname{} Reward Hit better reflects reasoning-path soundness.}
    \label{fig:correlation}
  \end{figure}

\begin{table*}[ht]
\centering
\caption{Results on Knights and Knaves dataset.}\label{tab:kk}
\small
\renewcommand{\arraystretch}{1.5} 
\begin{tabular}{llccccccc}
\toprule
Method & Metric (\%) & 2ppl & 3ppl & 4ppl & 5ppl & 6ppl & 7ppl & 8ppl \\
\midrule
\multirow{2}{*}{Outcome-CoT} 
 & Answer Correct Rate  & 99.63 & 99.81 & 99.81 & 98.13 & 98.69 & 95.13 & 94.25 \\
 & GPT Soundness  & 64.81 & 68.56 & 69.13 & 61.75 & 56.06 & 53.88 & 47.94 \\
\midrule
\multirow{2}{*}{PRoSFI} 
 & Answer Correct Rate & 99.75 & 99.31 & 99.75 & 99.63 & 99.25 & 98.56 & 97.38 \\
 & GPT Soundness & 
   \makecell{86.56 \\ {\scriptsize\color{blue}{$\uparrow 21.8$}}} &
   \makecell{90.31 \\ {\scriptsize\color{blue}{$\uparrow 21.8$}}} &
   \makecell{90.50 \\ {\scriptsize\color{blue}{$\uparrow 21.4$}}} &
   \makecell{90.44 \\ {\scriptsize\color{blue}{$\uparrow 28.7$}}} &
   \makecell{91.00 \\ {\scriptsize\color{blue}{$\uparrow 35.0$}}} &
   \makecell{86.31 \\ {\scriptsize\color{blue}{$\uparrow 32.4$}}} &
   \makecell{80.94 \\ {\scriptsize\color{blue}{$\uparrow 33.0$}}} \\
\bottomrule
\end{tabular}
\end{table*}

\subsection{\shortname{} Reward Aligns Well with Soundness}

To evaluate the effectiveness of the reward design used in \shortname{}, we conduct a correlation analysis among three metrics: Answer Correctness, \shortname{} Reward Hit, and GPT-Soundness of generated reasoning paths. This analysis is performed on the outputs of \shortname{}, and the results are in Figure~\ref{fig:correlation}.

The results reveal that the \shortname{} Reward Hit signal exhibits a high correlation with GPT-Soundness, much higher than the Answer Correctness. This suggests that our reward function is more aligned with step-by-step reasoning quality, thereby achieving our design goal. In contrast, answer correctness alone may fail to capture subtle flaws or inconsistencies in intermediate reasoning steps. These findings highlight that \shortname{}'s reward design provides meaningful supervision that better promotes logically coherent and sound outputs.

\section{\shortname{} Generalizes to Knights and Knaves Data}

To demonstrate the generalization ability of our proposed method, we extend experiments to the Knights and Knaves dataset~\cite{xie2024memorization}, whose details can be found in Appendix~\ref{app:dataKK}. 

For the Outcome-CoT baseline, we adopt the same prompt template as Logic-RL, while the prompt for \shortname{} is adapted slightly to suit this task. Following the out-of-distribution (OOD) evaluation protocol established in Logic-RL, our model is trained exclusively on instances with difficulty levels from 3ppl to 7ppl. Evaluation is then conducted across the full test range, from 2ppl to 8ppl, with the 8ppl split serving as the primary benchmark for OOD performance.



The training configuration closely follows that of ProverQA, including a batch size of 64, 8 sampled responses per prompt, a maximum sequence length of 5120, and the same KL coefficient and clip-higher strategy from DAPO, except that we set the learning rate to $4 \times 10^{-7}$, consistent with Logic-RL, and extend the training schedule to 1440 steps to align with our experimental design.

From the results shown in Table~\ref{tab:kk}, we can see that on the Knights and Knaves dataset, ProSFI consistently improves GPT soundness over Outcome-CoT across all difficulty levels, while maintaining comparable answer accuracy, demonstrating strong logical consistency and  reliable generalization, particularly on the hardest 8ppl split. Therefore, our approach is generalizable to other logical reasoning tasks beyond ProverQA.
\section{Conclusion} \label{sec:discussion}

It is known that LLMs may make mistakes in the reasoning steps even when the final answer is correct, which can undermine their credibility in high-risk domains. 
In this work, we study the problem of how  to enable large language models generating step-by-step verifiable reasoning paths via  formal prover's feedback. We propose a novel method \fullname{} that connects natural language reasoning with formal provers' verification. By training LLM to output structured formal intermediates and validating them with formal provers, we obtain precise, step-level soundness signals. These signals serve as stepwise formal rewards during reinforcement learning, enabling the model to learn verifiably correct reasoning behaviors.


While our current framework focuses on first-order logic tasks, future work can extend it to broader domains such as general natural language reasoning and automatic theorem proving. These extensions will require more diverse datasets, meticulously designed cold-start strategies, stronger base models, and deeper integration with domain-specific formal reasoning tools.

Another viable direction involves further exploration of training algorithm development. Instead of relying solely on binary pass/fail signals, future systems could incorporate more granular reward functions or more precise credit assignment mechanisms to enhance training efficacy and stability.

\bibliography{Xproof}
\bibliographystyle{icml2026}

\newpage
\appendix
\onecolumn
\newpage
\appendix
{\LARGE\textbf{Appendices}}

\lstset{
  breaklines=true,
  breakatwhitespace=true,
  postbreak=\mbox{\textcolor{gray}{\(\hookrightarrow\)}\space}
}








\section{Use of Large Language Models (LLMS)}
While preparing this manuscript, we made use of Large Language Models (LLMs) as writing aids, specifically for tasks such as grammar correction, wording refinement, and rephrasing to enhance clarity and readability. In line with ethical guidelines, we, the authors, confirm that we take full responsibility for the content of this work. Any text produced with the assistance of LLMs has been carefully reviewed, revised, and approved by us. All scientific claims, findings, and conclusions are entirely our own. We are responsible for any possible mistakes, inaccuracies, or ethical issues that may arise in this submission.

\section{Details of Datasets}\label{app:data}
\subsection{ProverQA dataset}
The ProverQA dataset contains 1,500 problems evenly split into easy, medium, and hard subsets, each defined by controlled complexity. 
For each difficulty level, we first filtered out the instances labeled as "Uncertain" and reserved 64 instances examples for testing. To assess generalization to even harder problems, we further synthesized 128 additional extra hard examples using the same framework, referred to as the \emph{Extra Split}.

Every instance of ProverQA comprises:

\begin{itemize}
    \item A background story created for each problem to introduce linguistic context and variation,
\item A logic skeleton of FOL premises (facts + rules with 7 operators: $\land, \lor, \lnot, \to, \oplus, \forall, \exists$) and distractors (non-essential facts or semantically related but irrelevant rules),
\item A goal to classify as \emph{True, False, or Uncertain},
\item Intermediate reasoning chains provided in both symbolic form and natural language. 
\end{itemize}

This design meets key criteria: scalability, natural \& diverse language, explicit symbolic representations, and faithful reasoning chains. At the same time, it is challenging enough, for example, GPT4o achieves accuracy 46.2  with standard prompts, and 50 with CoT prompts for hard samples~\citep{qi2025large}.   Hence it is good to work with ProverQA for logical reasoning studies.

\subsection{An Example from ProverQA}





\lstdefinestyle{mystyle}{
  frame=single,
  basicstyle=\ttfamily\footnotesize,
  keywordstyle=\color{blue},
  stringstyle=\color{darkred},
  numberstyle=\tiny\color{gray},
  emphstyle=\color{pink}\underbar,
  escapeinside={(*@}{@*)},
  breakatwhitespace=false,         
  breaklines=true,                 
  captionpos=b,                    
  keepspaces=true,                    
  numbersep=5pt,                  
  showspaces=false,                
  showstringspaces=false,
  showtabs=false,
  tabsize=2
}


\lstset{
    style=mystyle,
    basicstyle=\ttfamily\scriptsize,
    backgroundcolor=\color{white},
    stringstyle=\color{black},
    keywordstyle=\color{black},
    mathescape=true,
}

\definecolor{codegray}{rgb}{0.5,0.5,0.5}


\definecolor{codepurple}{rgb}{0.58,0,0.82}
\tcbset{
  promptbox/.style={
    colback=gray!10,    
    colframe=gray!50,   
    boxrule=0.5pt,      
    arc=2pt,            
    left=6pt,           
    right=6pt,          
    top=4pt,            
    bottom=4pt,         
    fonttitle=\bfseries,
    title={An Example from ProverQA}
  }
}

\begin{tcolorbox}[promptbox]
\begin{lstlisting}[language=python]
{
  "id": 0,
  "options": [
    "A) True",
    "B) False",
    "C) Uncertain"
  ],
  "answer": "B",
  "question": "Based on the above information, is the following statement true, false, or uncertain? Brecken has never experienced heartbreak.",
  "reasoning": "fact1: Brecken has experienced heartbreak.\nrule: Either Brecken has experienced heartbreak or he has never experienced heartbreak, but not both.\nconclusion: Brecken has experienced heartbreak.\n\nTherefore, it is false that Brecken has never experienced heartbreak. The correct option is: B.",
  "context": "Brecken has experienced heartbreak. Either Brecken has experienced heartbreak or he has never experienced heartbreak, but not both.",
  "nl2fol": {
    "Brecken has experienced heartbreak.": "has_experienced_heartbreak(Brecken)",
    "Either Brecken has experienced heartbreak or he has never experienced heartbreak, but not both.": "has_experienced_heartbreak(Brecken) (*@{{$\oplus$}}@*) has_never_experienced_heartbreak(Brecken)"
  },
  "conclusion_fol": "has_never_experienced_heartbreak(Brecken)"
}
\end{lstlisting}
\end{tcolorbox}

To illustrate the structure of the ProverQA dataset, we present a representative example as follows. Each entry is organized as a JSON object containing the question, multiple-choice options, the correct answer, and the reasoning chain used to derive the conclusion. In addition, the dataset provides a context field capturing the relevant facts and rules, as well as a mapping from natural language (NL) to first-order logic (FOL) formulas (nl2fol). Finally, the conclusion\_fol field specifies the logical statement under evaluation.

This example highlights how ProverQA combines natural language reasoning with its corresponding logical formalization, enabling the evaluation of models not only on final predictions but also on their formal reasoning steps.
\subsection{Knight and Knaves Dataset}\label{app:dataKK}

The Knights and Knaves (K\&K) puzzles\citep{xie2025logic} are an algorithmically generated reasoning dataset where characters are either knights, who always tell the truth, or knaves, who always lie. The goal is to determine each character’s role from their statements. The dataset is procedurally generated, providing unlimited variation and unseen test cases for evaluating generalization. Difficulty is precisely controlled by varying the number of characters (2–8) and the complexity of logical operations, supporting curriculum learning and out-of-distribution testing. Each puzzle has a single verifiable solution, ensuring evaluations rely on strict deductive reasoning.

K\&K dataset is particularly useful due to its formal, synthetic design. Every puzzle is governed by explicit rules and admits a unique, deterministic solution, eliminating ambiguity and allowing researchers to clearly separate true reasoning ability from memorization.

Here is an example adapted from the \cite{xie2025logic}.

\definecolor{codepurple}{rgb}{0.58,0,0.82}
\tcbset{
  promptbox/.style={
    colback=gray!10,    
    colframe=gray!50,   
    boxrule=0.5pt,      
    arc=2pt,            
    left=6pt,           
    right=6pt,          
    top=4pt,            
    bottom=4pt,         
    fonttitle=\bfseries,
    title={An Example from K\&K Puzzle Problem}
  }
}

\begin{tcolorbox}[promptbox]
\begin{lstlisting}[language=python]
(*@\textbf{Problem:}@*) A very special island is inhabited only by knights and knaves. Knights always tell the truth, and knaves always lie. You meet 2 inhabitants: Zoey, and Oliver. Zoey remarked, "Oliver is not a knight". Oliver stated, "Oliver is a knight if and only if Zoey is a knave". So who is a knight and who is a knave? 

(*@\textbf{Solution:}@*) (1) Zoey is a knave (2) Oliver is a knight

\end{lstlisting}
\end{tcolorbox}

\section{Algorithm Details}\label{app:algorithm}

For completeness, we revisit the GRPO algorithm in detail. The procedure consists of the following steps:
\begin{enumerate}
  \item Samples a group of output sequences \(\{o_1, \dots, o_G\}\) for each prompt \(q\) from the  policy \(\pi_{\theta_{\mathrm{old}}}\).
  \item Scores each sequence \(o_i\) with a reward model \(r_\phi(q, o_i)\), yielding rewards \(\{r_i\}_{i=1}^G\).
  \item Computes relative advantages by normalizing these rewards within the group:
  \[
    \hat A_{i} 
      = \frac{r_i - \tfrac{1}{G}\sum_{j=1}^G r_j}
             {\sqrt{\tfrac{1}{G}\sum_{j=1}^G \bigl(r_j - \tfrac{1}{G}\sum_{k=1}^G r_k\bigr)^2}}
    .
  \]
  \item Optimizes the policy by maximizing a PPO-style clipped surrogate objective, with a KL penalty against a frozen reference model \(\pi_{\mathrm{ref}}\).

\end{enumerate}

Moreover in the experiments, we implement the clip-higher strategy from the Decoupled Advantage Policy Optimization (DAPO) algorithm~\citep{yu2025dapo} to stablize the training of GRPO. This strategy increases the upper clip range of the importance sampling ratio in the policy gradient loss to encourage exploration. Consistent with the DAPO methodology, we set the asymmetric clipping range with $\varepsilon_{\text{low}}=0.2$ and $\varepsilon_{\text{high}}=0.28$.

\section{Prompts for The Outcome-CoT Approach} \label{app:OutcomePrompt}

\definecolor{codepurple}{rgb}{0.58,0,0.82}
\tcbset{
  promptbox/.style={
    colback=gray!10,    
    colframe=gray!50,   
    boxrule=0.5pt,      
    arc=2pt,            
    left=6pt,           
    right=6pt,          
    top=4pt,            
    bottom=4pt,         
    fonttitle=\bfseries,
    title={System Prompt for Outcome-CoT}
  }
}

\begin{tcolorbox}[promptbox]
\begin{lstlisting}[language=Python]
(*@\color{codepurple}{\textbf{System:}}@*)

Given a problem statement as contexts, the task is to answer a logical reasoning question.

The assistant first thinks about the reasoning process in the mind and then provides the user with the answer. 
The reasoning process is enclosed within <think> </think> tags, followed by the answer, i.e., <think> reasoning process here </think> answer here.

The last A), B), or C) in the answer section (i.e., the content outside <think> </think>) 
will be considered your answer. If the format is incorrect, you may not receive a score.
\end{lstlisting}
\end{tcolorbox}





The User Prompt for the Outcome-CoT approach used in Section \ref{sec:initial} is as follows.

The prompt   begins with a context section containing background knowledge expressed in natural language, including facts (e.g., Paola appreciates beauty) and rules (e.g., Anyone who can solve problems and communicate effectively has practical skills). These statements establish the logical environment within which reasoning must take place.

Following the context, the prompt specifies a question that requires truth evaluation of a target statement—in this case, whether Paola is artistically inclined. The prompt also provides a set of options (True, False, or Uncertain), mirroring the multiple-choice format used throughout the dataset. Finally, the template ends with the phrase “The correct option is:”, leaving space for the model to generate its answer.


\tcbset{
  promptbox/.style={
    colback=gray!10,    
    colframe=gray!50,   
    boxrule=0.5pt,      
    arc=2pt,            
    left=6pt,           
    right=6pt,          
    top=4pt,            
    bottom=4pt,         
    fonttitle=\bfseries,
    title={User Prompt for Outcome-CoT}
  }
}

\begin{tcolorbox}[promptbox]
\begin{lstlisting}[language=Python]
(*@\color{codepurple}{\textbf{User:}}@*)

(*@\textbf{Context:}@*)
Anyone who can solve problems and communicate effectively has practical skills.
If Paola appreciates beauty, then she values precision and enjoys manual work.
Paola appreciates beauty. 
Paola is either skilled in mathematics or has practical skills, but not both.
If someone enjoys manual work and values attention to detail, then they can craft with their hands.
Paola is either skilled in mathematics or artistically inclined, but surprisingly, she is actually both. 
Paola values attention to detail.
Anyone who crafts with their hands has practical skills.

$\textbf{Question:}$
Based on the above information, is the following statement true, false, or uncertain? 
Paola is artistically inclined.

$\textbf{Options:}$
A) True
B) False
C) Uncertain

The correct option is:
\end{lstlisting}
\end{tcolorbox}





\section{Prompts and Example Responses for \shortname{}}

\subsection{System Prompt for  \shortname{}}

To demonstrate how \shortname{} is guided in the reasoning process, we provide the system prompt used in Section~\ref{sec:grpo-json} (see Listing~X). The prompt instructs the model to behave as a mathematical reasoner that operates on premises expressed in Lean4 syntax (e.g., $\land$, $\lor$, $\lnot$, $\to$, $\oplus$, $\forall$, $\exists$).

The structure of the prompt has several key elements:
\begin{itemize}
    \item \textbf{Formatted reasoning}: All intermediate thinking is enclosed within \textless think\textgreater...\textless/think\textgreater.
    \item \textbf{Formatted summarized output}: Inside \textless summary\textgreater...\textless/summary\textgreater, the model must return a JSON array of atomic inference steps.
\end{itemize}

 Each \textbf{JSON object} specifies:
\begin{itemize}
    \item id: a unique identifier for the step,

\item dependencies: the premises or earlier steps it relies on,

\item conclusion: a single new FOL formula expressed in Lean4,

\item rule: the inference rule applied.
\end{itemize}

\textbf{Constraints} – Each inference step must be as fine-grained as possible (fewer than 5 dependencies), and the JSON must be valid and parsable. The final object corresponds to the model’s chosen answer and must align with one of the provided options.

This design enforces faithful, step-by-step reasoning, making the model’s decision process transparent and verifiable.

\tcbset{
  promptbox/.style={
    colback=gray!10,    
    colframe=gray!50,   
    boxrule=0.5pt,      
    arc=2pt,            
    left=6pt,           
    right=6pt,          
    top=4pt,            
    bottom=4pt,         
    fonttitle=\bfseries,
    title={System Prompt for PRoSFI}
  }
}

\begin{tcolorbox}[promptbox]
\begin{lstlisting}[language=Python]
(*@\color{codepurple}{\textbf{System:}}@*)

You are a math reasoner. For each problem, you will receive a set of premises annotated in Lean4 (use (*@{{$\to$}}@*), (*@{{$\oplus$}}@*), (*@{{$\land$}}@*), (*@{{$\lor$}}@*), (*@{{$\lnot$}}@*), (*@{{$\forall$}}@*), (*@{{$\exists$}}@*)). Answer the multiple-choice reasoning question as follows:

1. Internal reasoning goes inside <think>$\cdots$</think>.
2. Then, inside <summary>$\cdots$</summary>, emit a JSON array of atomic inferences, one per object:

  ```json
  {
    "id": "s",
    "dependencies": ["$h_1$","s"], // original premises or earlier steps
    "conclusion": "$\dots$",           // exactly one new FOL formula (Lean4 syntax)
    "rule": "$\dots$"                  // the single inference rule applied
  }
  ```
  
  - The JSON array must be valid and parsable without errors.
  - Each object must represent exactly one atomic inference step.
  - Each step should be as atomic as possible, using less than 5 dependencies.

The last object in the array corresponds to your answer. Its id and conclusion must be identical to one of the options.
\end{lstlisting}
\end{tcolorbox}

\subsection{User Prompt for PRoSFI}
To make the interaction format concrete, we show the \textbf{user prompt} used for \shortname{} in Section~\ref{sec:grpo-json}. The prompt is divided into several components:

\begin{itemize}
    \item \textbf{Context}: A sequence of natural language premises paired with their corresponding formal statements in Lean4 syntax. For example, the first two premises assert conditional rules about Hattie and Candy, while the 3rd and 4th connect moral courage to principled action.
\item \textbf{Question}: A query formulated in natural language, asking whether a target statement (here, “Hattie does not gain community respect”) is true or false given the context.

\item \textbf{Options}:  The possible answers, each expressed both in natural language and as a formal Lean4 statement.

\item \textbf{Answer slot}: The prompt ends with “The correct option is:”, leaving space for the model to provide its choice.
\end{itemize}

This structure illustrates how \shortname{} is guided to align natural language reasoning with formal logical inference, ensuring that both the problem and the candidate answers are clearly specified in parallel in both natural language and formal statement.

\tcbset{
  promptbox/.style={
    colback=gray!10,    
    colframe=gray!50,   
    boxrule=0.5pt,      
    arc=2pt,            
    left=6pt,           
    right=6pt,          
    top=4pt,            
    bottom=4pt,         
    fonttitle=\bfseries,
    title={User Prompt for PRoSFI}
  }
}

\begin{tcolorbox}[promptbox]
\begin{lstlisting}[language=Python]
(*@\color{codepurple}{\textbf{User:}}@*)

(*@\textbf{Context:}@*)
1. If Hattie stands up for her principles, then she gains community respect. Formal statement: `$h_1$ : stands_up_for_principles Hattie $\to$ gains_community_respect Hattie`.
2. If Candy stands up for her principles, then she gains community respect. Formal statement: `$h_2$ : stands_up_for_principles Candy $\to$ gains_community_respect Candy`.
3. Hattie has moral courage. Formal statement: `$h_3$ : has_moral_courage Hattie`.
4. Everyone who has moral courage will stand up for their principles. Formal statement: `$h_4$ : $\forall x$, (has_moral_courage $x \to$ stands_up_for_principles $x$)`.

Question: Based on the above information, is the following statement true or false? Hattie does not gain community respect.

Options:
A) True. Formal statement: `h_goal_true: $\lnot$gains_community_respect Hattie`.
B) False. Formal statement: `h_goal_false: $\lnot$($\lnot$gains_community_respect Hattie)`.

The correct option is:
\end{lstlisting}
\end{tcolorbox}


\subsection{Model Response after RL with PRoSFI}
We next illustrate how the model responds after  RL with PRoSFI.  The response is divided into two main parts:

\begin{itemize}
    \item \textbf{Reasoning path}: Enclosed within <think> … </think>, the model provides a step-by-step explanation in natural language. In this case, it applies Universal Instantiation and Modus Ponens to derive that Hattie stands up for her principles, and subsequently that she gains community respect. This reasoning shows why the negated claim (Hattie does not gain community respect) is false.
\item \textbf{Structured summary}: Enclosed within <summary> … </summary>, the model outputs a JSON array of atomic inference steps. Each object specifies the inference id, the set of dependencies, the derived conclusion (in Lean4 syntax), and the rule applied. The last object corresponds to the final answer and matches one of the candidate options.
\end{itemize}

We can see the model’s response is both interpretable in natural language and machine-verifiable through its structured reasoning steps.

\begin{lstlisting}[language=Python]
(*@\color{codepurple}{\textbf{Assistant:}}@*)

<think>
1. From $h_4$ ($\forall x$, has_moral_courage x $\to$ stands_up_for_principles x) and $h_3$ (has_moral_courage Hattie), by Universal Instantiation on Hattie and Modus Ponens, derive stands_up_for_principles Hattie.
2. From $h_1$ (stands_up_for_principles Hattie $\to$ gains_community_respect Hattie) and the above, by Modus Ponens, derive gains_community_respect Hattie.
3. Therefore $\lnot$ gains_community_respect Hattie is false.
</think>
<summary>[
  {
    "id": "s1",
    "dependencies": ["$h_3$", "$h_4$"],
    "conclusion": "stands_up_for_principles Hattie",
    "rule": "Universal Instantiation + Modus Ponens"
  },
  {
    "id": "h_goal_false",
    "dependencies": ["$h_1$", "s1"],
    "conclusion": "$\lnot$($\lnot$gains_community_respect Hattie)",
    "rule": "Modus Ponens"
  }
]
</summary>
\end{lstlisting}

\section{ Typical Errors for \texttt{Outcome-CoT}}
\label{sec:informal-error}

In this section, we select two representative cases to analyze two typical types of errors that occur during the reasoning process of the \texttt{Outcome-CoT} model, namely semantic confusion and commonsense shortcuts.

\subsection{Semantic Confusion}

\begin{lstlisting}[language=Python]
(*@\color{codepurple}{\textbf{User:}}@*)

(*@\textbf{Context:}@*)
(*@$\cdots$@*)
3. Either Miracle is community-oriented or self-centered, but not both.
(*@$\cdots$@*)
6. If Miracle is a leader, then she is either selfless or thinks of herself, but not both.
(*@$\cdots$@*)
(*@\color{codepurple}{\textbf{Assistant:}}@*)
(*@$\cdots$@*)
Since she is community-oriented, she cannot be self-centered (statement 3). Therefore, she must be selfless. Thus, the statement 'Miracle is not self-centered' is true.
(*@$\cdots$@*)
\end{lstlisting}


In this example, the model infers that ``\texttt{Miracle is not self-centered}'' from the premise that ``\texttt{Miracle is community-oriented}'', and then equates ``\texttt{not self-centered}'' with ``\texttt{selfless}''. In doing so, it bypasses a rigorous logical analysis of the premises and directly concludes that ``\texttt{Miracle is selfless}''. However, determining whether Miracle is truly \texttt{selfless} requires a more comprehensive evaluation based on Proposition 6, whether Miracle is a leader, as well as the mutually exclusive conditions specified by the two choices. Thus, the model’s conclusion reflects a logical error arising from semantic confusion.


\subsection{Commonsense Shortcut}

\begin{lstlisting}[language=Python]
(*@\color{codepurple}{\textbf{User:}}@*)

(*@\textbf{Context:}@*)
(*@$\cdots$@*)
6. If Alvaro faces adversity, then he either stays motivated or pursues his goals, but not both.
(*@$\cdots$@*)
(*@\color{codepurple}{\textbf{Assistant:}}@*)
(*@$\cdots$@*)
Since Alvaro is facing adversity and is either staying motivated or pursuing his goals, it is clear that Alvaro is overcoming challenges in some form. Therefore, the statement "Alvaro does not overcome challenges" is false.
(*@$\cdots$@*)
\end{lstlisting}


In this example, after confirming that Alvaro is facing adversity and satisfies either ``\texttt{staying motivated}'' or ``\texttt{pursuing goals}'', the model proceeds to conclude that ``\texttt{Alvaro overcame the challenge}''. Although this inference may seem reasonable from a commonsense perspective, it lacks a rigorous logical basis. The context of the question does not provide information on whether Alvaro actually overcame the challenge. Therefore, the model’s conclusion is a flawed logical generalization based on a commonsense shortcut, which fails to adhere to the principles of deductive reasoning.

\section{Details and Credibility Analysis of the GPT Soundness Metric}

\begin{tcolorbox}[systempromptbox]
\begin{lstlisting}[language=Python]
(*@\color{codepurple}{\textbf{System:}}@*)

The following is a problem statement and a reasoning chain to solve it.

(*@\textbf{Problem:}@*)
...

(*@\textbf{Reasoning:}@*)
...

(*@\textbf{Task:}@*)
Please evaluate whether the reasoning process contains any irreparable logical errors.
- If so, return a JSON where "is_error" is True, "error_instance" contains the original text of the first instance of the error, "error_type" specifies the type of the error, and "error_reason" explains the reason for the error.
- Otherwise, return a JSON where "is_error" is False.
Wrap your JSON within ```json ```.
\end{lstlisting}
\end{tcolorbox}





We use a simple prompt to ask the GPT-OSS-120B model to evaluate the correctness of the reasoning path and ask it to return a JSON; when the reasoning is correct, it only needs to return a bool value, but when it is incorrect, it needs to provide the location and reason for the error.

We manually checked 10 outputs each from \texttt{Outcome-CoT} and \shortname{} to evaluate the correctness of GPT's assessment of the reasoning process.

For the outputs of \texttt{Outcome-CoT}, GPT's evaluations were all correct. Among them, 8 outputs had errors of invalid deduction, one output was completely correct, and one output had an error of misinterpretation of premise.

For the outputs of \shortname{}, GPT's evaluations were almost entirely correct but slightly too strict. Among them, 5 outputs had errors of invalid deduction, while 3 outputs were completely correct. The remaining two outputs had reasoning that was generally correct but contained some minor errors: the first output used unnecessary premises in one step, and the second output wrote the wrong name for the reasoning rule used.

\section{Reward Function for \shortname{} Step-by-Step Verification}

We first extract a structured JSON representation from the natural language reasoning paths generated by the model. We then determine whether the format is correct based on its validity and the structural specifications defined in the main text. By analyzing the \texttt{id} 
 and \texttt{conclusion}  fields of the last object in the JSON, we can determine whether the final answer is correct.

To verify the logical correctness of the entire reasoning process, we use a top-down syntax parser to parse the first-order logic (FOL) expressions generated by the model. These are then converted into formats supported by target formal tools such as Lean4, Prover9, or Z3. For each reasoning conclusion, we express it as a proposition in Lean4 and verify it using a sequence of tactics or a large language model capable of formal theorem proving. Alternatively, the reasoning steps can be converted into input formats accepted by Prover9 or Z3, and the validity of each step can be verified using an automated solver. 



\begin{algorithm}
\caption{Reward Function for \shortname{} Step-by-Step Verification}
\label{algo:algorithm1}
\begin{algorithmic}[1]
\STATE \textbf{Input:} response, ground\_truth
\STATE \textbf{Output:} score

\STATE summary $\leftarrow$ \texttt{extract\_summary(response)}
\STATE statements $\leftarrow$ \texttt{extract\_fol\_problems(summary, ground\_truth)}

\IF{statements is None}
   \STATE \textbf{return} 0.0
\ENDIF

\IF{summary.last\_id $\ne$ ground\_truth}
   \STATE \textbf{return} \texttt{format\_score}
\ENDIF

\IF{\texttt{fol\_all\_solved(statements)}}
   \STATE \textbf{return} \texttt{full\_score}
\ELSE
   \STATE \textbf{return} \texttt{option\_score}
\ENDIF
\end{algorithmic}
\end{algorithm}

The algorithm is presented in Algorithm~\ref{algo:algorithm1}, which follows this routine:
\begin{enumerate}
    \item  Extract a summary and FOL statements from the model's response.
\item If no valid statements are extracted, return a score of 0.
\item If the ID of the final reasoning step doesn't match the ground truth, return a score reflecting formatting issues.
\item  If all FOL problems are solved correctly, return the full score.
\item  Otherwise, return a partial (optional) score.

\end{enumerate}





\section{Correlation Analysis of the Results on \texttt{Outcome-CoT}}



\begin{figure*}[htb]
\centering
\includegraphics[width=0.8\linewidth]{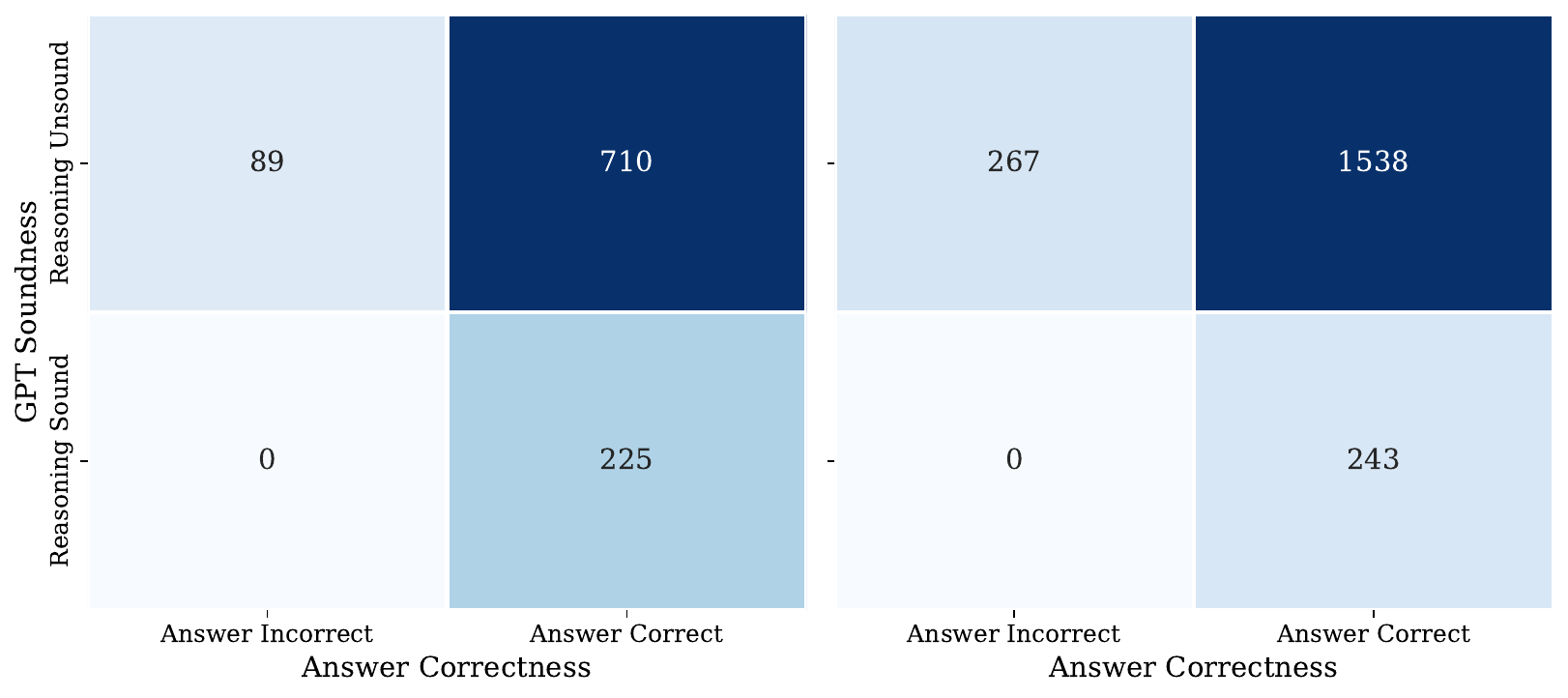}
\caption{\textbf{Left:} Confusion matrix of answer correctness vs. reasoning soundness for \texttt{Outcome-CoT} on the \texttt{ProverQA-hard} subset.
\textbf{Right:} Corresponding results on the \texttt{ProverQA-extra} subset. The lack of alignment between correct answers and sound reasoning indicates that answer accuracy alone is not a sufficient criterion for judging the quality of reasoning.}
\label{fig:confusion}
\end{figure*}

For the output of the \texttt{Outcome-CoT} method, we analyze the relationship between two available metrics: the correctness of the final answer and the reasoning soundness automatically evaluated by GPT (GPT Soundness). The confusion matrix is shown in Figure~\ref{fig:confusion}.

It can be observed that a large number of reasoning paths, although yielding the correct answer, are evaluated as unsound in their reasoning process. This result indicates that a correct answer alone is not sufficient to reflect the logical validity of the reasoning process. It further supports the qualitative observation in Section~\ref{sec:initial} — namely, that traditional CoT methods struggle to ensure the validity of reasoning paths.

In comparison, Figure~\ref{fig:correlation} presents the heatmaps of variable correlations on the two subsets. The results show that the designed reward signal has a stronger correlation with GPT Soundness than with the final answer correctness. This suggests that our formal verification mechanism effectively identifies and enhances logical rigor, making it a more reliable signal for training and selection.

\end{document}